\documentclass[journal]{IEEEtran}

\usepackage[pdftex]{graphicx}
\ifCLASSINFOpdf
\else
\fi
\usepackage{verbatim}
\usepackage{amsmath}
\usepackage{soul}

\usepackage{color}
\usepackage{multirow}

\begin{document}
%
\title{Sequence-to-Sequence Load Disaggregation Using Multi-Scale Residual Neural Network}
%
%
%

\author{Gan~Zhou,~\IEEEmembership{Member,~IEEE,}
		Zhi~Li,
        Meng~Fu,
        Yanjun~Feng,
        Xingyao~Wang
        and~Chengwei~Huang,~\IEEEmembership{Member,~IEEE}
\thanks{Manuscript received April 19, 2005; revised August 26, 2015. This work was partially supported by the National Natural Science Foundation of China (Grant No. 51877038).}
\thanks{Gan Zhou (corresponding author), Zhi Li, Meng Fu and Yanjun Feng are with the School of Electrical Engineering, Southeast University, Nanjing, China e-mail: \{zhougan2002,lizhi95,fumeng,fengyanjun\}@seu.edu.cn.}
\thanks{Xingyao Wang is with the School of Instrument Science and Engineering, Southeast University, Nanjing, China e-mail: 230189843@seu.edu.cn.}
\thanks{Chengwei Huang is with Jiangsu Intever Energy Technology Co. Ltd., Nanjing, China e-mail: \{huangchengwei\}@intelever.com.}
}

%
%

\markboth{Journal of \LaTeX\ Class Files,~Vol.~14, No.~8, August~2015}%
{Shell \MakeLowercase{\textit{et al.}}: Bare Demo of IEEEtran.cls for IEEE Journals}
%



\maketitle

\begin{abstract}
With the increased demand on economy and efficiency of measurement technology, Non-Intrusive Load Monitoring (NILM) has received more and more attention as a cost-effective way to monitor electricity and provide feedback to users. Deep neural networks has been shown a great potential in the field of load disaggregation. In this paper, firstly, a new convolutional model based on residual blocks is proposed to avoid the degradation problem which traditional networks more or less suffer from when network layers are increased in order to learn more complex features. Secondly, we propose dilated convolution to curtail the excessive quantity of model parameters and obtain bigger receptive field, and multi-scale structure to learn mixed data features in a more targeted way. Thirdly, we give details about generating training and test set under certain rules. Finally, the algorithm is tested on real-house public dataset, UK-DALE, with three existing neural networks. The results are compared and analysed, the proposed model shows improvements on F1 score, MAE as well as model complexity across different appliances.
\end{abstract}

\begin{IEEEkeywords}
Non-intrusive load monitoring, residual neural network, multi-scale, model complexity, load disaggregation.
\end{IEEEkeywords}

%
\IEEEpeerreviewmaketitle

\section{Introduction}
With the increasingly complicated demands of household electricity consumption, the emergence of smart grids and load monitoring is indispensable. In this field, measurement technology plays a rather important role. Obtaining useful information usually requires the correct installation, maintenance and interpretation of a large number of sensors. Even if the sensors have been mass-produced with low-cost, this is a difficult task \cite{Shaw}. Therefore, how to deploy load monitoring through little measurement cost and faster measurement methods is an urgent problem to be solved.

Load monitoring includes Intrusive Load Monitoring (ILM) and Non-Intrusive Load Monitoring (NILM). ILM refers to the installation of sensors on each of users' electrical appliances. Though the monitoring data obtained in this way is accurate and reliable, there are still disadvantanges of large investment, poor operability and low user acceptance \cite{Ridi}. NILM, which acquires only one measurement point at the user's house entrance, is a more economical and effective method. It disaggregates the user's total load by analysing the load features, and estimates the power consumption of a single electric appliance \cite{Hart}.

Existing methods for NILM in the literature include the feature-based approach \cite{alacala, Chang_new, Yi_new, Rahimi_new, Su_new} and the data-driven approach \cite{Jiang_new, Liu_Akintayo, Nalmpantis_new}. The former does not require statistical learning of a model, but only signal-level processing \cite{Aboulian, Dinesh} is needed to achieve optimal results. The latter relies on a certain amount of labelled data \cite{Kelly2} to learn a representative model. Data-driven methods can further be roughly classified into conventional machine learning \cite{KongW}, deep learning \cite{LiuY} and online transfer learning \cite{Iwayemi}.

Feature-based methods depend on the correct detection of ``on-and-off'' events of an appliance. Signal processing techniques are used to improve the features which are key to event detection. Electric features like current, voltage and active power are the waveform based time-domain features. On the other hand, advanced non-stationary signal processing algorithms bring more transformation domain features, such as spectral features \cite{Dinesh} and wavelet based features \cite{Tabatabaei,Gillis}.

With enough ``hand-crafted'' feature descriptors, optimization framework can be applied to these non-training methods \cite{Machlev}. However, the generalization ability always relies on the ``hand-crafted'' expert knowledge. This may be a drawback in some large-scale deployment of energy management systems.

The data-driven methods adopt statistic learning frameworks, such as single-label classification, multi-label classification \cite{Tabatabaei}. Among the deep neural network approaches \cite{S_Ding, Liang_new}, the depth of the network continues to increase in order to achieve a better performance. Meanwhile, the network faces the problem of degradation, which includes gradient explosion and gradient vanishing. Although some of the existing models can achieve good disaggregation performance under a mixture of simple appliances, the disaggregation performance in a complex case is not satisfactory.

In this paper, we propose a novel neural network model for blind load disaggregation: a multi-scale deep residual neural network based on dilated convolution. This model outperforms several existing deep neural networks on load disaggregation and it shows a considerable improvement on model complexity.

The structure of this paper is as follows: Sec.\ref{relatedwork} provides the related work and reviews the previous NILM methods. Sec.\ref{sec_method} introduces the proposed neural network model structure. Sec.\ref{sec_exp} introduces the generation of training set and test set, the experimental settings and results. A further comparison of various neural network models is also provided. In Sec.\ref{sec_conclusion} we provide the conclusion of this work and plans for future work.

\section{Related Work}
\label{relatedwork}

The conventional methods are based on event detection \cite{Hart}. The Factor Hidden Markov Model (FHMM) proposed by Kim et al. \cite{Kim, Palma} outperformed some basic statistical models. However, it still has some limitation when adapted to complex real-world situations. Neural network, on the other hand, has a strong  non-linear modeling ability given sufficient quantity of data. Roos et al. applied neural network model in NILM field for the first time \cite{ROOS}, though the model lacked for learning capacity due to simple structure. More recent development on deep learning has provided us various efficient models, mostly Long-Short Term Memory Recurrent Neural Networks (LSTM-RNN), denoising autoencoders (dAE), and Convolutional Neural Networks (CNN).

Mauch and Yang \cite{Yang} proposed to use a RNN network with simply connecting multiple LSTM layers together. Aiming differently, Bongli et al. constructed their network to be a denoising autoencoder, which considers the aggregate power as a noisy version of appliance load, and applied a median filter on the output \cite{Bongli}. Similarly, Zhang at al. adopted CNN but more in a conventional structure, where fully-connected layers follow a succession of CNN layers. They applied their model on a sequence-to-sequence situation \cite{ZhangZhong}, where the output is the disaggregated sequence, as well as a sequence-to-point one \cite{Incecco}, where the output is a single load of a specified time.

In order to benchmark the deep neural networks on load disaggregation, P. Nascimento performed experiment on various typical RNNs included Simple Recurrent Network (SRN), LSTM and GRU, CNN and Recurrent Convolutional Network (RCN) with different activation functions \cite{Nascimento}. They used three types of common appliances as the disaggregation objects. Kelly and Knottenbelt \cite{Kelly} proposed three different models including BiLSTM, dAE and a regression network. In their work, the former two networks performed sequence-to-sequence disaggregation and the third performed sequence-to-point disaggregation, where the output was the start time, end time and the average power consumption within the input window. Furthermore, He and Chai \cite{ChaiW} proposed a BiLSTM and a dAE with parallel convolution structure.

Beyond simple traditional network structures, Harell et al. adopted Time Convolutional Network (TCN) with dilated convolution \cite{Harell}. They retained the causality of the input sequences and used gated structure to extract complex features. Baets et al. explored a Siamese network structure consisted of two CNNs with V-I trajectory of a high sample frequency as the input \cite{Baets}.

Although most of the works exhibited above still used active power as the sole input feature because it is simple to measure and existing metering infrastructures usually provide the necessary values \cite{pochacker}. 
D. Srinivasan proposed a network using harmonic source as the input \cite{Srinivasan}, and Rosdi proposed a Probabilistic Neural Network (PNN) using magnetic field waveforms as input \cite{Rosdi}.

An open challenge in this field is the adaptation ability. There are many different makes and models, and labelled training datasets are not always enough. Therefore, Yanchi Liu et al. \cite{LiuY} used transfer learning to make use of a visual neural network in appliance identification. Similarly, semi-supervised learning \cite{Iwayemi,Gillis} and some empirical techniques may be also used to train the model on one site and deploy on another.

\section{Methodology}
\label{sec_method}
\begin{figure}[!t]
\centering
\includegraphics[width=3.5in]{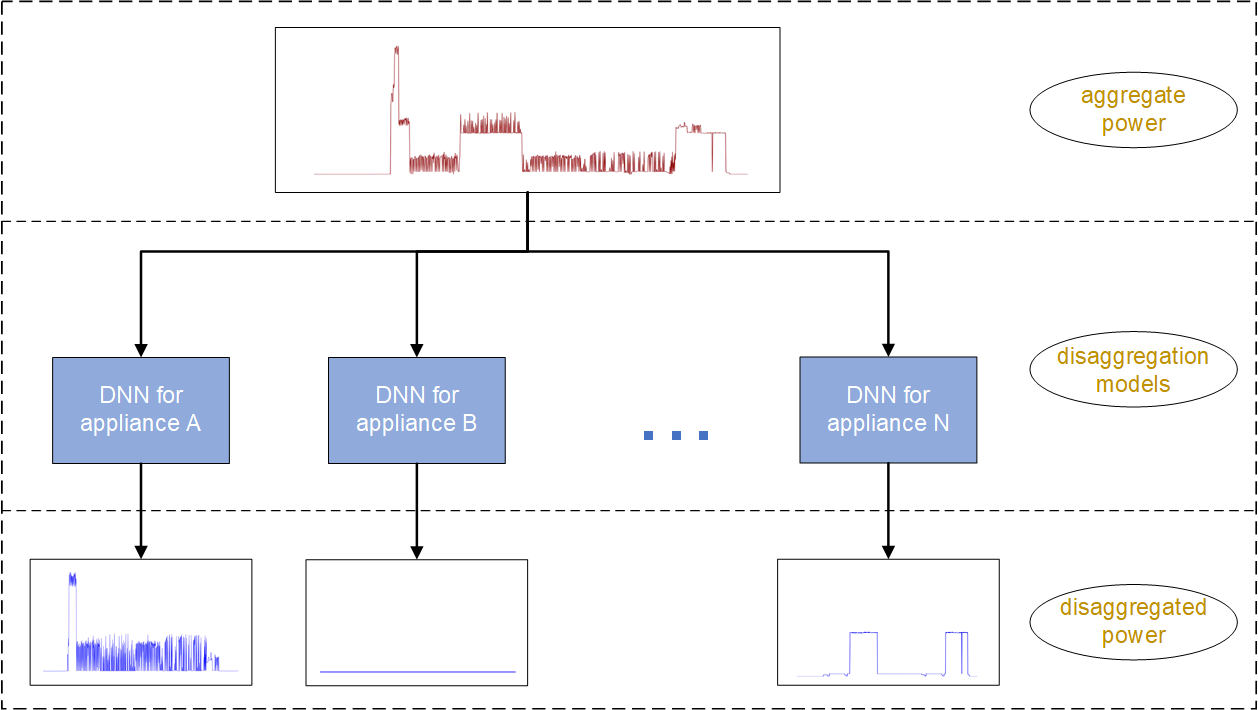}
\caption{Process of the source separation for NILM}
\label{fig.seq2seq}
\end{figure}

In order to separate the active power data of each appliance respectively from the aggregate active power data. In this paper, we are inspired by Kelly et al. \cite{Kelly} and train a neural network for each type of appliances; the inputs to the neural network are aggregate power sequences, and the network accordingly outputs sequences of equal length that contains only the active power consumption of a single appliance. Its general process is shown in Fig.\ref{fig.seq2seq}.

\subsection{Dilated Convolutional Residual Block}
\label{sec_Residual}
In this paper, we propose a residual network architecture \cite{HeZhang} which is consisted of dilated convolutional residual blocks as basic structural units. Residual blocks have a considerable advantage in avoiding gradient vanishing or explosion problem when the network layers increases. 
Multi-branches designed in the proposed network tend to learn representations from different receptive fields parallelly with less information loss.

\begin{figure}[!t]
\centering
\includegraphics[width=2.5in]{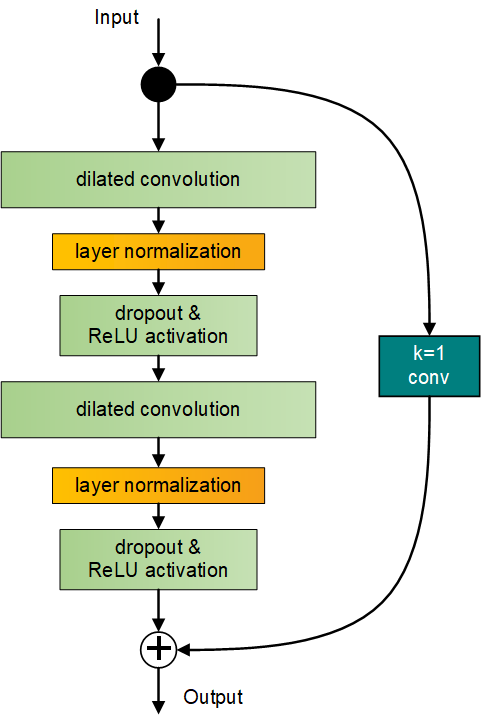}
\caption{Structure of a residual block}
\label{fig.block}
\end{figure}

The structure of the dilated convolutional residual block is shown in Fig.\ref{fig.block}. The input for each residual block successively goes through two convolutional layers and their corresponding layer normalizations, dropouts and nonlinear activations. Then the original input will be added to the temporal output by using a shortcut connection, together the two constitute the final output of this residual block. If the number of input channels is different from that of the temporal output channels, a convolutional layer of kernel size of 1 will be adopted to alter the number of channels to be equal to the temporal output's before adding them up.

The shortcut connection enables each residual block to learn features directly from the original input, which greatly improves the learning capacity of the network. We connect these residual blocks one after another to form a residual block body, in order to extend the depth of residual network.


The dilated convolution in the residual block has two other features compared to conventional convolution. First, due to the characteristics of the sequence-to-sequence, we need to ensure that the sequences before and after each convolution have the same length; second, the kernels will have holes (i.e., zero values) in them whose number depends on the dilation rate of that layer, a feature that allows the network to obtain a larger receptive field by selectively skipping some part of the inputs without changing the kernel sizes or increasing parameters. If $d$ denotes the dilation rate, then there will be $d-1$ values skipped between every two actual values convoluted by the kernel. When $d=1$, it is obvious that the dilated convolution reverts to an ordinary one-dimensional convolution.

\begin{figure}[!t]
\centering
\includegraphics[width=3.5in]{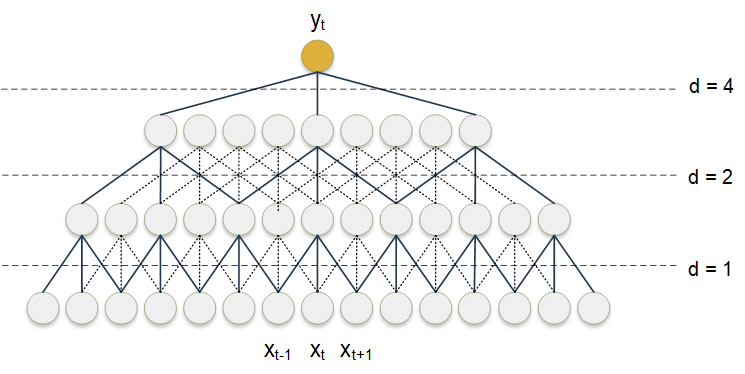}
\caption{Non-causal dilated convolution}
\label{fig.cnn}
\end{figure}

In the proposed network, both convolutional layers in each residual block have the same dilation rate, and the dilation rate of the residual block will gradually increase exponentially as the network gets deeper, as shown by the simplified schematic of the dilated convolutional network in Fig.\ref{fig.cnn} where $d$ denotes dilation rate, and the receptive field gets larger in the meantime. The figure shows the situation where kernel size is 3.

$X=\{x_0,x_1,\dots,x_{T-1},x_T\}$ is the original input sequence, $Y=\{y_0,y_1,\dots,y_{T-1},y_T\}$ is the final output sequence. It is clear that for the timing sequences, the convolution has a feature of non-causality \cite{Bai}, meaning the output at time $t$ is determined by an equal number of values both from past and future. Load disaggregation can also be seen as a denoising problem, so more information from the sample context can help the network to learn better about the features of the pure signal. In order to achieve the above non-causality, each layer of zero padding needs to be set symmetrically to both ends of the feature map. Given $k$ denotes kernel size and $d$ denotes dilation rate, the number of zero paddings on both ends is $(k-1) \times d \div 2$.

For LSTM, each time step calculation must wait until the previous one is completed in order to maintain the time-complexity of the sequence, which leads to much consuming of training and test time. In contrast with LSTM, the dilated convolutional network has the advantage of processing multiple time steps in parallel, and also of the flexibility of the receptive field. In particular, when the dilation rate of the last residual block is $2^{D}$, the receptive field can be calculated as in Eq.\ref{receptive_field}

\begin{equation}
\begin{split}
S&=(2 \times \sum_{d=0}^D \frac{k-1}{2} \times 2^{d+1})+1\\
&=(2^{D+2}-2)(k-1)+1\label{receptive_field}
\end{split}
\end{equation}
where $k$ denotes the kernel size (in case of an odd number) and $d$ denotes the dilation rates in the residual block body.

\subsection{Load Disaggregation Model}
\label{sec_nilm}

\begin{figure*}[!t]
\centering
\includegraphics[width=5.0in]{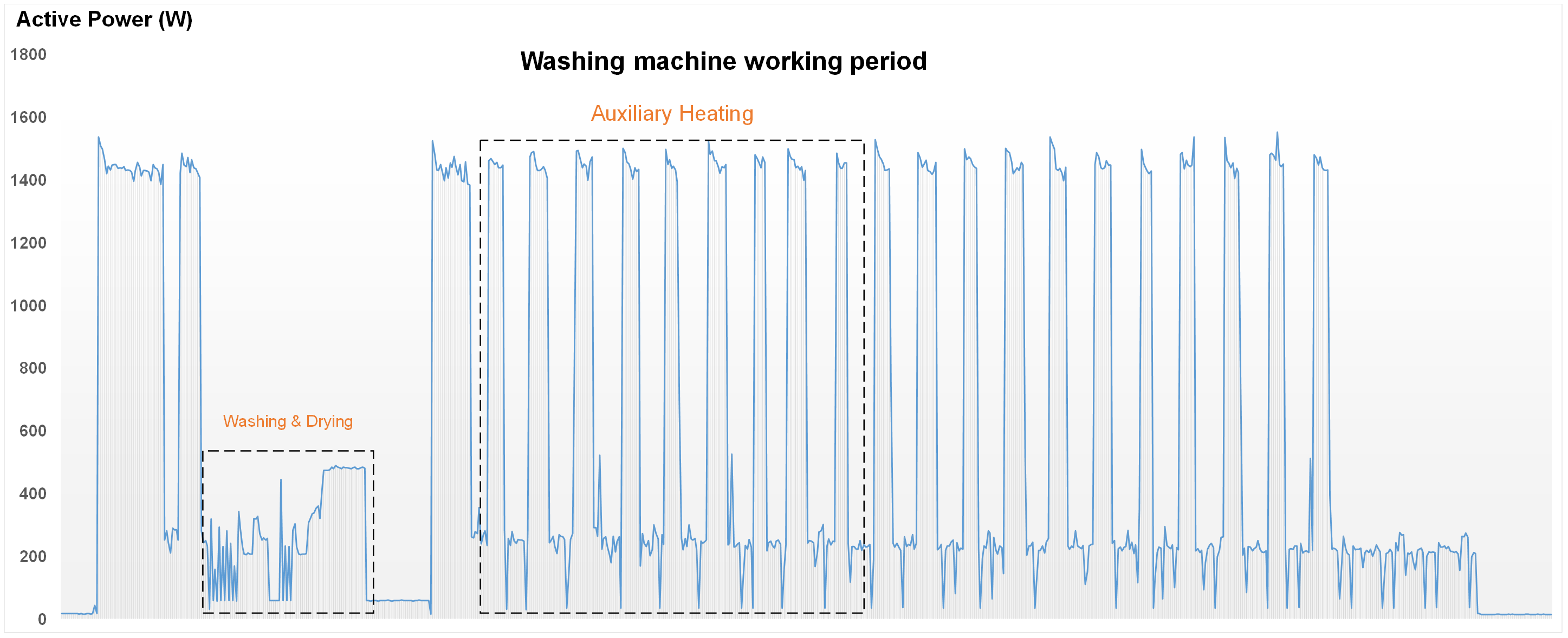}
\caption{Power profile of a working period of washing machine}
\label{fig.working}
\end{figure*}

Considering that different appliances have different working modes, even for the same appliances, there are different sub-modes in a working cycle. In these different working modes and sub-modes, the features of the power profile, including duration, power, and fluctuations are all different from each other. As an example, the active power profile of the washing machine is shown in Fig.\ref{fig.working}. The working modes in the two dashed boxes are washing and auxiliary heating, which have obvious differences in power features.

If a receptive field can only cover a particular working mode, the network's representation of this specific feature mode is better. But if it's blindly increased, the learning may be aimless, and the features of different working modes may be mixed and overlapped with each other, which will worsen the final disaggregation performance.

\begin{figure*}[!t]
\centering
\includegraphics[width=5.5in]{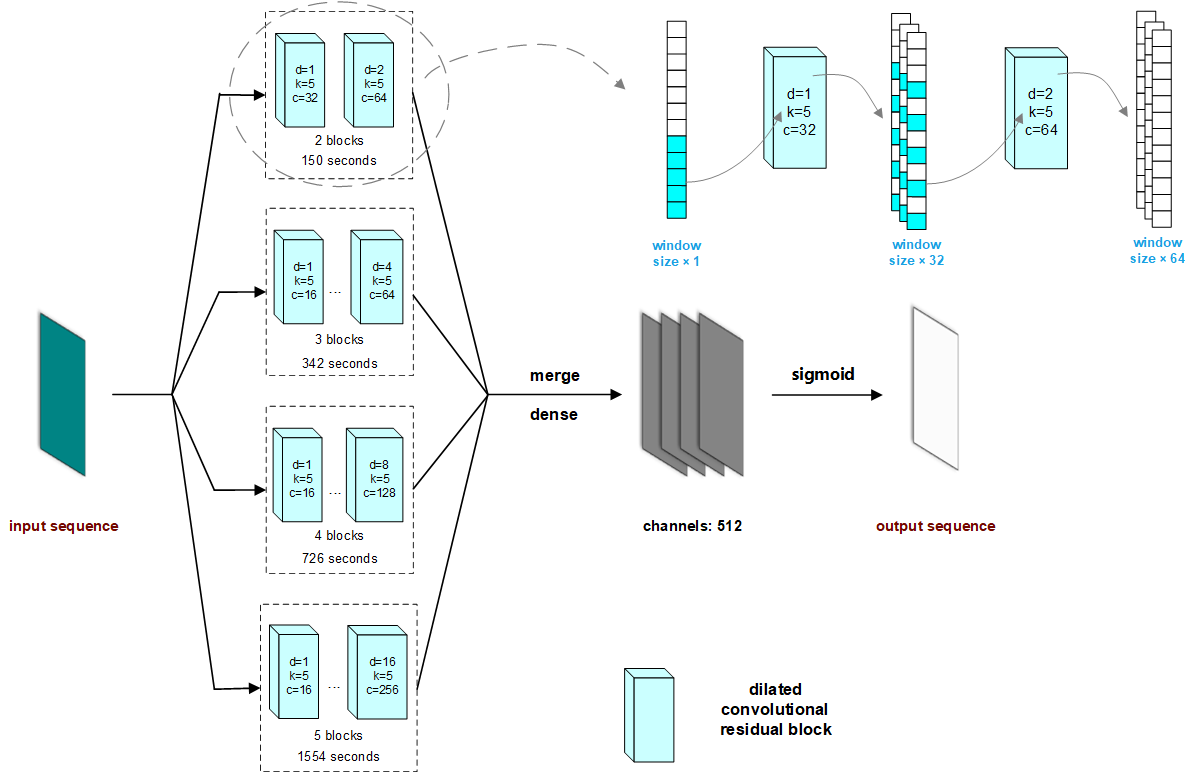}
\caption{Overall structure of the multi-scale residual network based on dilated convolution}
\label{fig.arch}
\end{figure*}

Taking this into consideration, we design a novel structure in such a way: merging the different receptive fields together, 
so that network itself is able to learn representations from multiple receptive fields combined with activation duration of different appliance.
The proposed \textit{multi-scale residual network} structure is shown in Fig.\ref{fig.arch}, where $c$ denotes output channels, $d$ denotes dilation rate and $k$ denotes kernel size.

As shown in Fig.\ref{fig.arch}, first, the input sequence is sent to four residual block bodies with a kernel size of 5 that contain 2 blocks, 3 blocks, 4 blocks and 5 blocks to perform feature extraction. In every residual block body, the dilation rates and the output channels of the first and the last block are indicated. All others in between multiply successively.

Based on the equation \ref{receptive_field}, the receptive field can be calculated as respectively 25, 57, 121 and 259 sample points. The dataset we employ on experiment, UK-DALE, has a sample frequency of 6 seconds a point, so we can correspond the receptive field sizes with real-time durations of 150 seconds, 342 seconds, 726 seconds and 1554 seconds. Then, the outputs of each residual block body are concatenated on depth, sent to two fully connected layers, and output with sigmoid activation function. The activation functions of all the other layers are all ReLU. The optimizer we use in training is the Adam optimizer because its adaptiveness can accelerate the convergence of network; the loss function we use is the cross-entropy, for after compared with the mean squared error (MSE), cross-entropy shows much better performance.

We will test the proposed model's performance in \ref{sec_exp}.

\section{Experiments and Results}
\label{sec_exp}
In this section, we detail the dataset we use and how to generate the training and test set required for the experiment under certain rules. Then we give the structures of the three exisiting models, BiLSTM and dAE in \cite{Kelly} as well as conventional convolutional network in \cite{ZhangZhong}. They are compared with the \textit{multi-scale residual network} introduced in the previous section. Finally we present the metrics for evaluation of the experiment and discuss the results of the experiments.

\subsection{Dataset}

In our experiments, the dataset we employ is the public available dataset UK Domestic Application Level Electric (UK-DALE) to perform the experiment. UK-DALE records 5 houses, all of the appliance-level power data is active power sampled every 6 seconds, while the aggregate power data is a mixture of apparent power sampled every 6 seconds and active as well as reactive power sampled every 1 second. We down-sample all the active power to 6 seconds resolution and only select apparent power when there's no active power available.

\subsection{Data Pre-processing and Model Training}

The UK-DALE dataset has various of appliances of varying power levels to choose from, but for the sake of sample size, proportion of electricity consumption, and the representativeness of the power features, we choose the following five types of appliances to to carry out experiments: kettle, fridge, washing machine, microwave and dish washer. All five of these types are distributed in UK-DALE in at least three houses, and the exact distribution of each appliance can be seen in Tab.\ref{tab.apphouse}. We use the data in House 5 as the test set and the data in the other houses as the training set. In addition, by the fact that the washing machine in House 4 shares the same meter with the microwave, the power data of washing machine and microwave in House 4 will not be involved in training.

\begin{table}[!t]
\renewcommand{\arraystretch}{1.3}
\caption{The distribution of different appliances data in UK-DALE}
\label{tab.apphouse}
\centering
\begin{tabular}{|c|c|}
\hline
Appliance &	House ID\\
\hline
Kettle	& 1,2,3,4,5 \\
Fridge	& 1,2,4,5 \\
Washing machine	& 1,2,4,5 \\
Microwave	& 1,2,4,5 \\
Dish washer	& 1,2,5 \\
\hline
\end{tabular}
\end{table}

In order to provide training samples and labels, we need to first locate the ``on" period of an appliance. We use NILMTK's $Electric.get\_activation()$ function and the arguments provided by UK-DALE to obtain the activations. For each time index, there's a pair of aggregate power sequence $N={n_0,n_1,\dots,n_{t-1},n_{t}}$ and label power sequence $M={m_0,m_1,\dots,m_{t-1},m_t}$, and because the sequence lengths are different between pairs, we uniformly set a constant window length for each appliance. Then the training and test samples are generated by randomly placing the window over the activations at the precondition that the whole activation is contained.

The arguments for $Electric.get\_activation()$ and window lengths are shown in Tab.\ref{tab.activation} and Tab.\ref{tab.winlen}.


\begin{table*}[!t]
\renewcommand{\arraystretch}{1.3}
\caption{Arguments passed to $get\_activation()$}
\label{tab.activation}
\centering
\begin{tabular}{|c||c|c|c|}
\hline
Appliance &	On power threshold (W) & Min. on duration (s) &  Min. off duration (s) \\
\hline
Kettle &	2000	& 12	& 0 \\
Microwave	& 200	& 12	& 30  \\
Fridge &	50	 	& 60	& 12 \\
Dish washer	& 10	& 1800	& 1800 \\
Washing machine	& 	20	& 1800	& 160 \\
\hline
\end{tabular}
\end{table*}

\begin{table}[!t]
\renewcommand{\arraystretch}{1.3}
\caption{Window lengths used for different appliances}
\label{tab.winlen}
\centering
\begin{tabular}{|c|c|}
\hline
Appliance &	Length (num of points)\\
\hline
Kettle & 64 \\
Microwave	& 128 \\
Fridge	& 512 \\
Dish washer	& 1024 \\
Washing machine &	1024 \\
\hline
\end{tabular}
\end{table}

The input sequences of the network are segmented from the aggregated data, and the label sequences are segmented from the appliance-level data.

Before starting to train the model, we performed the following pre-processing on the dataset:

\begin{itemize}
  \item [1)] 
  We normalized every pair of training samples and labels in order to facilitate the training of network, that was, all divided by the maximum power value of the aggregate sequence in each pair, so that all values fell within the interval [0,1]. We also apply this rule when generating test set.    
  \item [2)]
  We put windows one window length ahead from the start point of activations to segment some amount of training samples that didn't contain activations of the target appliance, thus generalize the learning capacity of the network. We also apply this rule when generating test set.
  \item [3)]
  We filtered the training samples. The samples where the activation length was shorter than 1/3 of the window length, and the samples where the sum of points of which aggregate power was three times greater than appliance power exceeded half of the window length would be discarded and not be included in the training set. Noted that when generating test set, we will not apply this rule in order to obtain the most general experiment results.
\end{itemize}

After the training set was generated, we perform experiments with a conventional convolutional network, a dAE, a BiLSTM, and the proposed model and compare the disaggregation performance. We use optimizers and loss functions in accordance with original papers of the methods, for conventional convolutional network, we use Adam optimizer \cite{ZhangZhong}, and for dAE and BiLSTM, we use stochastic gradient descent (SGD) with Nesterov momentum of 0.9 \cite{Kelly}. All three of these models are trained with MSE as loss function.

\subsection{Comparison with Existing Network Structures}
Three exisiting neural network models \cite{Kelly, ZhangZhong} are adopted for comparison. They performed effectively in previous work on UK-DALE dataset.

\begin{figure*}[!t]
\centering
\includegraphics[width=5.5in]{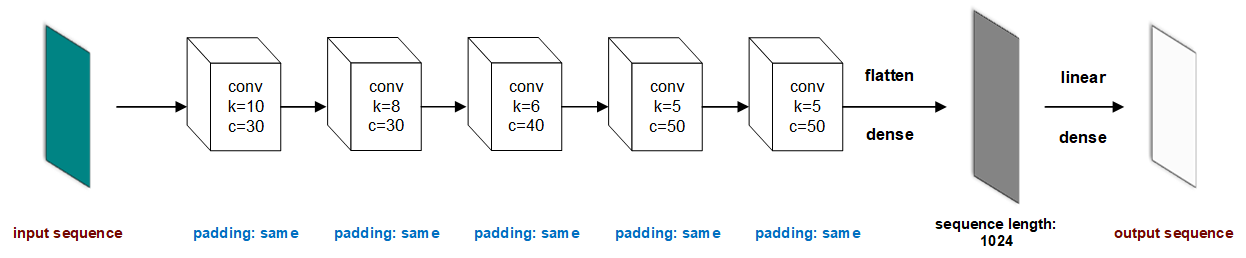}
\caption{Structure of conventional convolutional network}
\label{fig.conv}
\end{figure*}

The structure of conventional convolutional network is shown in Fig.\ref{fig.conv}.

\begin{figure*}[!t]
\centering
\includegraphics[width=5.5in]{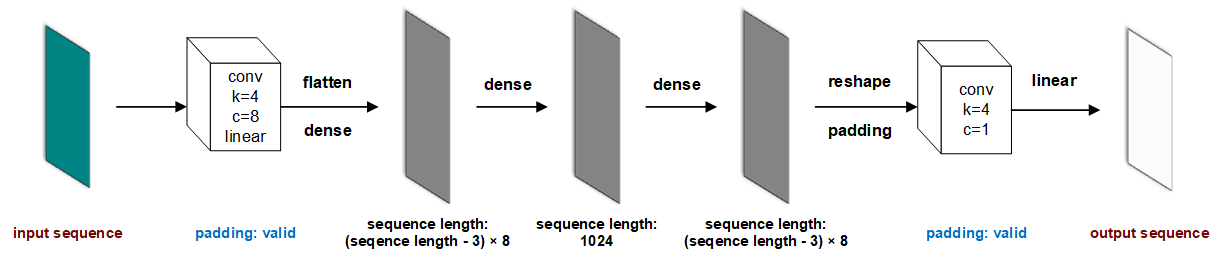}
\caption{Structure of denoising autoencoder}
\label{fig.dae}
\end{figure*}

The structure of dAE is shown in Fig.\ref{fig.dae}.

\begin{figure*}[!t]
\centering
\includegraphics[width=5.5in]{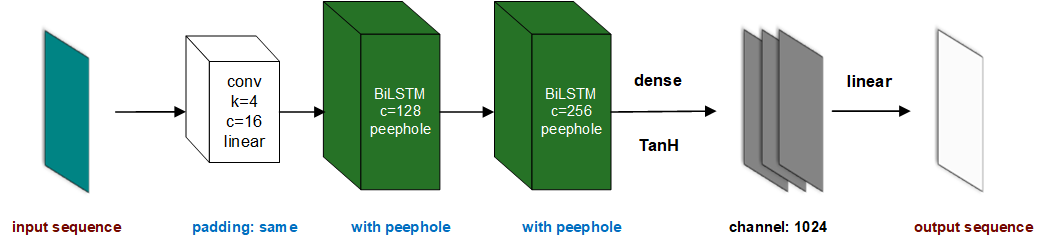}
\caption{Structure of BiLSTM}
\label{fig.bilstm}
\end{figure*}

The structure of BiLSTM is shown in Fig.\ref{fig.bilstm}.

\subsection{Experimental Results}
The metrics for evaluation of disaggregation performance are defined as below.

TP: true positive

FP: false positive

TN: true negative

FN: false negative

The actual power of an appliance at time $t$ is denoted as $y_t$ and the estimated power is denoted as $\hat{y}_t$

Four types of static metrics are defined as:

\begin{equation}
Recall = \frac{TP}{TP+FN}
\end{equation}

\begin{equation}
Precision = \frac{TP}{TP+FP}
\end{equation}

\begin{equation}
F1 Score = 2 \times \frac{Precision \times Recall}{Precision + Recall}
\end{equation}

\begin{equation}
Mean Absolute Error = \frac{\sum_{t=1}^T |y_t-\hat{y}_t|}{T}
\end{equation}

Recall and precision are often considered as intermediate metrics, for achieving high score on either can still possibly leading to poor F1 score, which reflects the performance of the network as a whole in the aspect of classifying the ``on-and-off" state of the appliances. Mean absolute error (MAE), on the other hand, reflects the performance in terms of more specific power deviations.

\begin{figure*}[!t]
\centering
\includegraphics[width=6in]{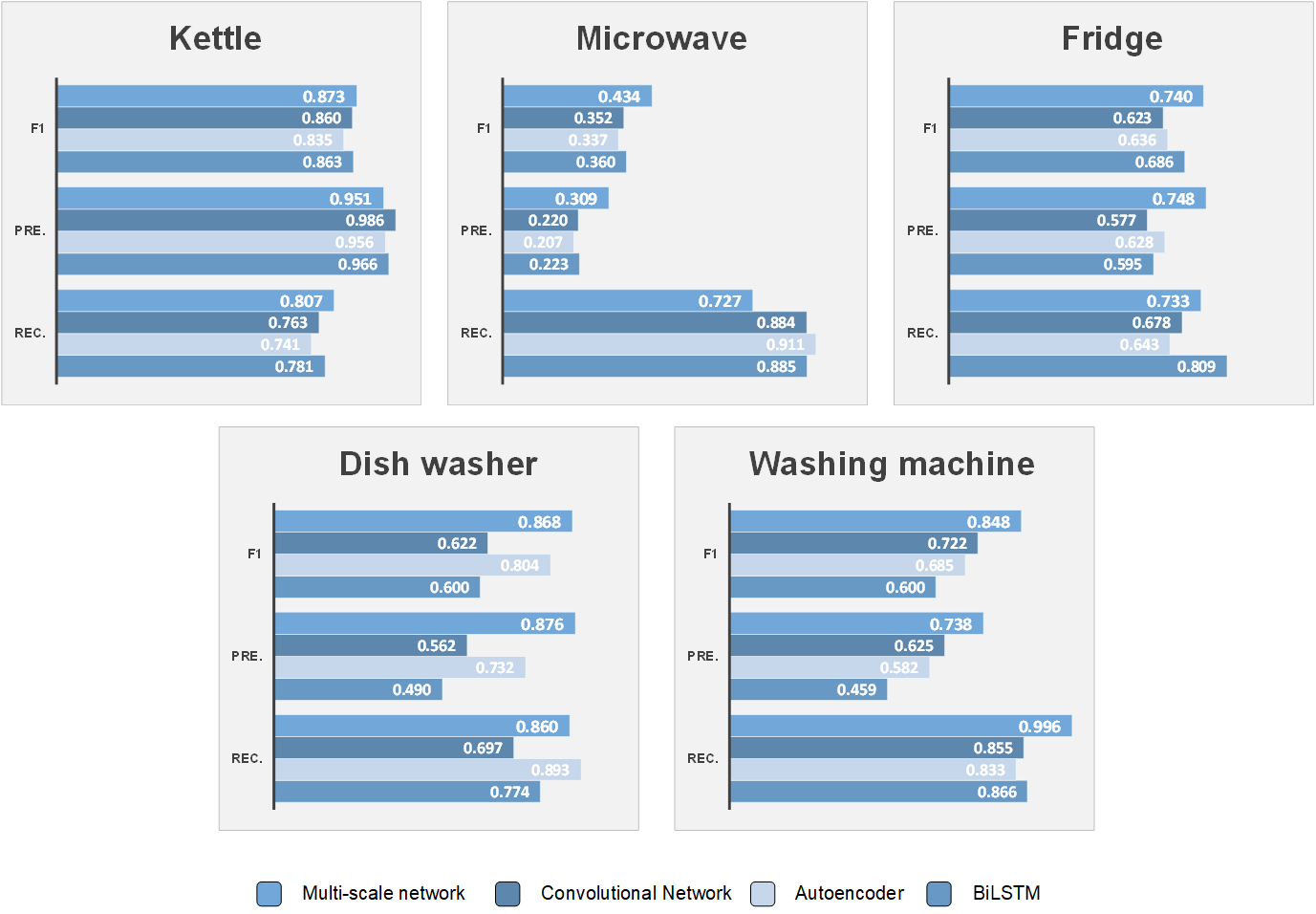}
\caption{F1, recall and precision results}
\label{fig.f1results}
\end{figure*}

\begin{figure}[!t]
\centering
\includegraphics[width=3.5in]{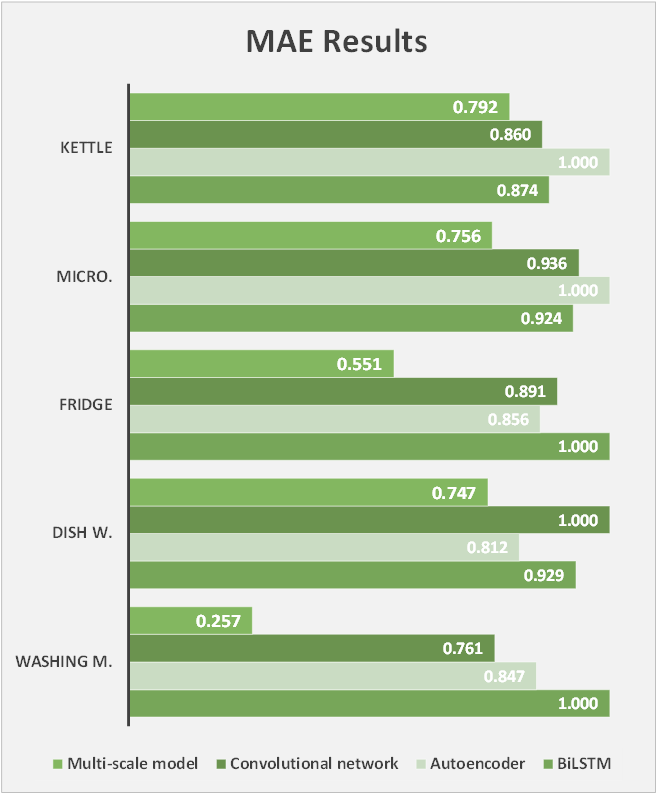}
\caption{Normalized MAE results}
\label{fig.maeresult}
\end{figure}

The experiment is performed in an environment described in Tab.\ref{tab.environment} and the testing results on different types of appliances using different models are shown in Fig.\ref{fig.f1results} and Fig.\ref{fig.maeresult}. We use Tensorflow Profiler to add up the number of parameters for each model and time.time() function to measure the prediction time for each sample. The models' parameters and time results are shown in Tab.\ref{tab.param} and Tab.\ref{tab.costtime}.

\begin{table}[!t]
\renewcommand{\arraystretch}{1.3}
\caption{Experimental environment}
\label{tab.environment}
\centering
\begin{tabular}{|c|c|}
\hline
CPU & GPU \\
\hline
Intel ® Core ™ i9-9900k 3.60GHZ &	NVIDIA RTX TITAN 24GB \\
\hline
\end{tabular}
\end{table}

All MAE results of each appliance are divided by the maximum MAE value of that appliance so that they can be normalized within [0,1] and distinctly exhibited in figure. The denominators are shown in Tab.\ref{tab.mae}.

\begin{table*}[!t]
\renewcommand{\arraystretch}{1.3}
\caption{Statistics of parameters for each model}
\label{tab.param}
\centering
\begin{tabular}{|c|c|c|c|c|}
\hline
\multirow{2}{*}{Appliance} & \multicolumn{4}{c|}{Parameters}                                                                                                   \\ \cline{2-5} 
                           & CNN                     & \multicolumn{1}{l|}{Autoencoder} & \multicolumn{1}{l|}{BiLSTM} & \multicolumn{1}{l|}{Multi-scale model} \\ \hline
Kettle                     & 3.38M                   & 0.36M                             & \multirow{5}{*}{1.27M}      & \multirow{5}{*}{1.22M}                 \\
Microwave                  & 6.72M                   & 1.26M                            &                             &                                        \\
Fridge                     & 26.78M                  & 17.63M                           &                             &                                        \\
Dish washer                & \multirow{2}{*}{53.52M} & \multirow{2}{*}{68.82M}          &                             &                                        \\
Washing machine            &                         &                                  &                             &                                        \\ \hline
\end{tabular}
\end{table*}

\begin{table*}[!t]
\renewcommand{\arraystretch}{1.3}
\caption{Prediction time per sample}
\label{tab.costtime}
\centering
\begin{tabular}{|c|c|c|c|c|}
\hline
\multirow{2}{*}{Appliance} & \multicolumn{4}{c|}{Time (ms)}                                                                                  \\ \cline{2-5} 
                           & CNN   & \multicolumn{1}{l|}{Autoencoder} & \multicolumn{1}{l|}{BiLSTM} & \multicolumn{1}{l|}{Multi-scale model} \\ \hline
Kettle                     & 0.940 & 0.708                            & 78.1                        & 5.83                                   \\
Microwave                  & 1.06  & 0.714                            & 154                         & 5.92                                   \\
Fridge                     & 1.17  & 0.832                            & 623                         & 6.19                                   \\
Dish washer                & 1.37  & 1.23                             & 1205                        & 6.44                                   \\
Washing machine            & 1.37  & 1.23                             & 1236                       & 6.65                                   \\ \hline
\end{tabular}
\end{table*}

\begin{table*}[!t]
\renewcommand{\arraystretch}{1.3}
\caption{Maximum MAE values}
\label{tab.mae}
\centering
\begin{tabular}{|c|c|}
\hline
Appliance & Maximum MAE\\
\hline
Kettle &	289.044 \\
Microwave	& 322.362 \\
Fridge &	71.388 \\
Dish washer	& 158.11 \\
Washing machine &	217.517 \\
\hline
\end{tabular}
\end{table*}

From the results, it can be seen that appliances with simple power features, such as kettle and dish washer, can have better disaggregation performance than others, while appliances with features that are more difficult to capture, such as microwave, generally have poorer disaggregation performance.

Broken down to specific metrics, kettle has the best F1 scores on five models, but the MAEs are poor. Because kettle's activation is more concentrated within a small-sized window and its power level is higher (above 2000W), small errors in the neural network's output activated by sigmoid are amplified to very high levels. Likewise, models' better MAE performance on fridge, which has a lower power level, confirms the previous arguments. Washing machine and dish washer have lower power thresholds, so there are fewer false negative or false positive points, which resulted in a generally greater recalls than precisions; while the more comprehensive metrics such as F1s and MAEs differ.

Generally, the conventional convolutional network outperforms the autoencoder on both the kettle and the washing machine, but not the dish washer. BiLSTM achieves mediocre performance, with big MAEs and occasionally good F1 scores on three appliances. Meanwhile, the proposed multi-scale model achieves better performance on all five appliances.

Comparing the residual network based on dilated convolution with other CNN-based networks, whose F1 and MAE results of the five appliances are surpassed by that of proposed model, it can be seen that the residual network and dilated convolution can solve the problem of insufficient receptive field and degradation of the ordinary convolution.

As for the other parts, except for kettle, multi-scale models have the least parameters hence the smallest sizes. It enables already trained models to be more easily implanted in house-end measurement devices, hence a more convenient way to realize NILM methods in real-house situation.

When it comes to time cost, it can be seen that multi-scale residual network outperforms BiLSTM but not two other convolutional networks. It's because in spite of multi-scale model's edge on sizes, the more layers and convolutional operations lead to more floating point operations (FLOPs). On the other hand, BiLSTM's delay of prediction can be attributed to its lack of parallelism, every time step calculation must wait until all calculations of the last step are completed. So the parallel-processing feature of convolutional networks generally shortens the time cost, hence obtain better performance in real-house situation.

In summary, our proposed model has a faster and better disaggregation performance than RNNs, and it can also show advantages over other CNN-based models, with curtailing the number of parameters, and in the meantime not significantly increasing the time cost.

\section{Conclusion}
\label{sec_conclusion}
Based on dilated convolution, we propose a multi-scale residual neural network model for load disaggregation. We compare our model with three promising deep neural network models, a conventional convolutional network, a BiLSTM and a denoising autoencoder on the public dataset UK-DALE. We adopt F1 score and MAE as the metrics for evaluating the disaggregation performance.

The experimental results show that the dilated convolution and multi-scale structure can be successfully applied to time series load disaggregation, hence NILM problems. Compared with models based on ordinary convolution, the large receptive field generated by dilated convolution can obtain more information from the background to assist the disaggregation, and the multi-scale structure enables the model to achieve optimal performance when facing appliances with complex power features.

Furthermore, our proposed model shows a promissing potential on small size and is easy to implement in real-house situation. It also expedites the prediction time compared to BiLSTM with parallel-processing feature of CNN, although the actual FLOPs still drag a little behind shallower CNN networks.

In future work, we will extend the proposed model to sequence-to-point regression in order to directly estimate the disaggregated energy consumption.


%



\section*{Acknowledgment}
The authors would like to thank the anonymous reviewers for their valuable advices that improved this paper. This work was partially supported by the National Natural Science Foundation of China (Grant No. 51877038).

\ifCLASSOPTIONcaptionsoff
  \newpage
\fi



%

%

\begin{IEEEbiography}
[{\includegraphics[width=1in,height=1.25in,clip,keepaspectratio]{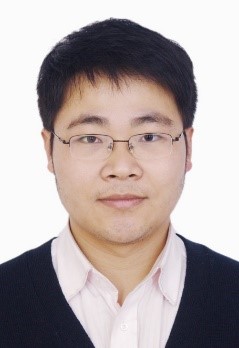}}]{Gan Zhou}

(M'09) received the M.S. and Ph.D. degrees from the School of Electrical Engineering, Southeast University, Nanjing, China, in 2003 and 2009, respectively. He is currently an Associate Professor with the School of Electrical Engineering, Southeast University. He has authored and co-authored over 40 papers in refereed journals and conference proceedings. His current research interests include nonintrusive load monitoring, big data analysis and high performance computing in power system.

\end{IEEEbiography}

\begin{IEEEbiography}
[{\includegraphics[width=1in,height=1.25in,clip,keepaspectratio]{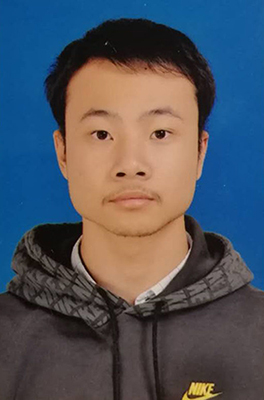}}]{Zhi Li}
received the B.S. degree in mechanical engineering from the Wuhan University of Technology, Wuhan, China, in 2017. He is currently pursuing M.S. degree from Southeast University, Nanjing, China. His current research interests include load disaggregation and monitoring, machine learning and data mining.
\end{IEEEbiography}

\begin{IEEEbiography}
[{\includegraphics[width=1in,height=1.25in,clip,keepaspectratio]{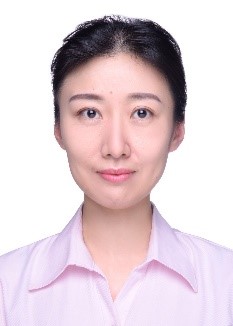}}]{Meng Fu}
received the B.S. and M.S. degree from the School of Electrical Engineering, Southeast University, Nanjing, China, in 2004 and 2007, respectively. She is currently pursuing Ph.D. degree from Southeast University. Her current research interests include power system analysis, deep learning and machine learning.
\end{IEEEbiography}

\begin{IEEEbiography}
[{\includegraphics[width=1in,height=1.25in,clip,keepaspectratio]{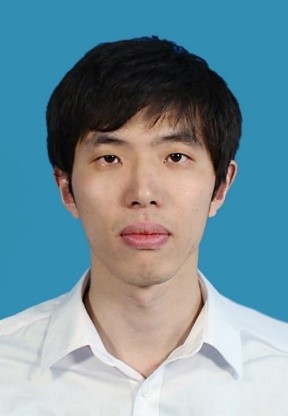}}]{Yanjun Feng}
received the B.S. degree in electrical engineering from the University of Electronic Science and Technology of China, Chengdu, China, in 2013. He received the M.S. degree in 2016 and is currently pursuing Ph.D. degree from Southeast University, Nanjing, China. His current research interests include load disaggregation and monitoring, data mining and GPU parallel computing in power system.
\end{IEEEbiography}

\begin{IEEEbiography}[{\includegraphics[width=1in,height=1.25in,clip,keepaspectratio]{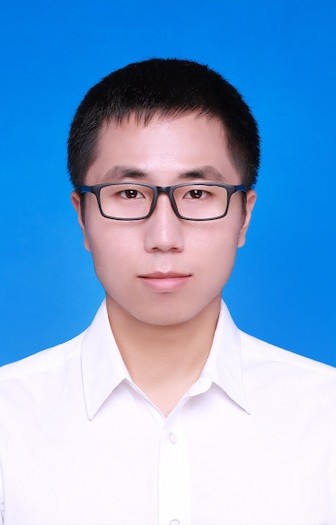}}]
{Xingyao Wang}
received his master degree in Signal Processing from Southeast University (SEU) in the P.R. China, 2018. Currently, he is working towards his Ph.D. degree in School of Instrument Science and Engineering, SEU. His research interests mainly lie in semantic segmentation, attention mechanism, causal system and deep learning for temporal data.
\end{IEEEbiography}

\begin{IEEEbiography}[{\includegraphics[width=1in,height=1.25in,clip,keepaspectratio]{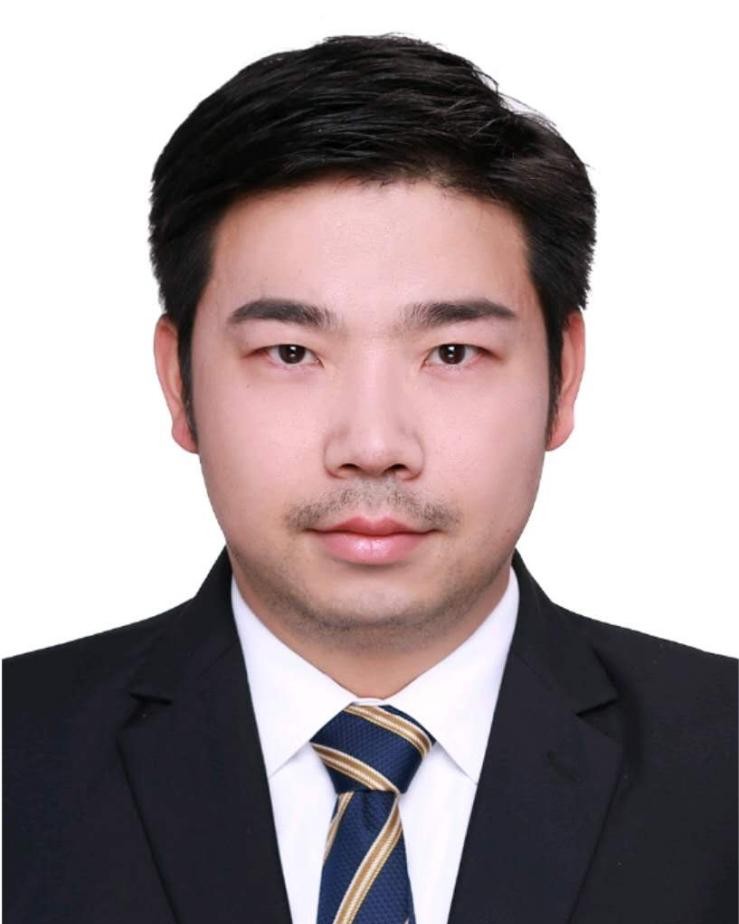}}]{Chengwei Huang}
(M'18) received his undergraduate degree in 2006, and Ph.D. for speech emotion recognition from Southeast University (China) in 2013. His main research interests include affective computing, signal processing and data mining.

He conducted research on human-computer interaction as an associate professor in Soochow University, and on big data technologies as CTO of Sugon (Nanjing) Institute of Chinese Academy of Sciences. He is currently with Intever Energy Technology Co. Ltd. focusing on AI industrial applications.
\end{IEEEbiography}




\end{document}